\def\year{2020}\relax
\documentclass[letterpaper]{article} % DO NOT CHANGE THIS
\usepackage{aaai20}  % DO NOT CHANGE THIS
\usepackage{times}  % DO NOT CHANGE THIS
\usepackage{helvet} % DO NOT CHANGE THIS
\usepackage{courier}  % DO NOT CHANGE THIS
\usepackage[hyphens]{url}  % DO NOT CHANGE THIS
\usepackage{graphicx} % DO NOT CHANGE THIS
\usepackage{graphicx}  % DO NOT CHANGE THIS
\usepackage{amsfonts}  % blackboard math symbols
\usepackage{amsmath}
\usepackage{amssymb}
\usepackage{amsthm}
\usepackage{appendix}
\usepackage{bbold}
\usepackage{booktabs}  % professional-quality tables
\usepackage{color}
\usepackage{courier}
\usepackage{float}
\usepackage{graphicx}
\usepackage{helvet}
\usepackage{mathtools}
\usepackage{mdwlist}  % single-spaced itemize* and enumerate*
\usepackage{microtype}  % microtypography
\usepackage{multirow}
\usepackage{natbib}
\usepackage{nicefrac}  % compact symbols for 1/2, etc.
\usepackage[T1]{fontenc}  % use 8-bit T1 fonts
\usepackage{tabularx}
\usepackage{times}
\usepackage{url}  % simple URL typesetting
\usepackage[utf8]{inputenc}  % allow utf-8 input
\usepackage{verbatim}
\usepackage{wrapfig}
\usepackage{xspace}

\usepackage{algorithm}
\usepackage{algorithmic}

\usepackage{hyperref}  % hyperlinks

\usepackage{subcaption}
\hypersetup{colorlinks = false, pdfborder = {0 0 0}}

\def\BibTeX{{\rm B\kern-.05em{\sc i\kern-.025em b}\kern-.08emT\kern-.1667em\lower.7ex\hbox{E}\kern-.125emX}}

\newif\ifdraft
\drafttrue

\ifdefined\isaccepted

\else

\fi

\renewcommand{\paragraph}[1]{\textbf{#1}\;\xspace}
\renewcommand{\eqref}[1]{(\ref{eq:#1})}

\newcommand{\figref}[1]{Figure~\ref{fig:#1}}
\newcommand{\tabref}[1]{Table~\ref{tab:#1}}

\newcommand{\eqrefs}[3][and]{(\ref{eq:#2}) #1~(\ref{eq:#3})}

\newcommand{\eqrefss}[4][and]{(\ref{eq:#2}), (\ref{eq:#3}) #1~(\ref{eq:#4})}

\newcommand{\eqrefsss}[5][and]{(\ref{eq:#2}), (\ref{eq:#3}), (\ref{eq:#4}) #1~(\ref{eq:#5})}

\newtheorem{theorem}{Theorem}

\newenvironment{titled-paragraph}[1]{\textbf{#1:}}{}

\newcommand{\iid}{\emph{i.i.d.}\xspace}

\newcommand{\ie}{i.e.\xspace}
\newcommand{\eg}{e.g.\xspace}
\newcommand{\egcite}[1]{\citep[\eg][]{#1}}

\newcommand{\probability}{\ensuremath{P}}
\newcommand{\expectation}{\ensuremath{\mathbb{E}}}

\newcommand{\indicator}{\ensuremath{\mathbb{I}}}

\newcommand{\defeq}{\ensuremath{\vcentcolon=}}

\newcommand{\N}{\ensuremath{\mathbb{N}}}

\newcommand{\R}{\ensuremath{\mathbb{R}}}

\newcommand{\maximize}[1][]{\ensuremath{\underset{#1}{\mathrm{maximize}}\;}}
\newcommand{\suchthat}[1][]{\ensuremath{\underset{#1}{\mathrm{s.t.}}\;}}

\newcounter{lineno}

\newcommand{\pacc}[2]{\ensuremath{A_{#1>#2}}}
\newcommand{\cacc}[1]{\ensuremath{A_{#1}}}

\newcommand{\cF}{\mathcal{F}}
\newcommand{\cX}{\mathcal{X}}

\newcommand{\cS}{\mathcal{S}}

\newcommand{\hlgt}[1]{{\color{blue} #1}}
\newcommand{\cL}{\mathcal{L}}

\title{Pairwise Fairness for Ranking and Regression}
\author{
Harikrishna Narasimhan, Andrew Cotter, Maya Gupta, Serena Wang\\
Google Research\\
1600 Amphitheatre Pkwy, Mountain View, CA 94043\\
 \{hnarasimhan, acotter, mayagupta, 
 serenawang\}@google.com 
}

\begin{document}

\maketitle

\begin{abstract}
We present pairwise fairness metrics for ranking models and regression models that form analogues of statistical fairness notions such as equal opportunity, equal accuracy, and statistical parity. Our pairwise formulation supports both discrete protected groups, and  continuous protected attributes. We show that the resulting training problems can be efficiently and effectively solved using existing constrained optimization and robust optimization techniques developed for fair classification. Experiments illustrate the broad applicability and trade-offs of these methods. 
\end{abstract}

\section{Introduction}\label{sec:introduction}
As ranking models and regression models become more prevalent and have a greater impact on people's day-to-day lives, it is important that we develop better tools to quantify, measure, track, and improve fairness metrics for such models. % Algorithms that respect fairness metrics are relatively well-studied for binary classification with categorical protected groups. Here, we extend the relatively mature work on training fairer binary classifiers to train fairer ranking and regression models. % To do so we re-frame the problem as comparing only pairs of examples, and this also empowers fairer ways to handle protected inputs that are continuous. 
A key question for ranking and regression is how to define fairness metrics. As in the binary classification setting, we believe there is not one ``right'' fairness definition: instead, we provide a paradigm that makes it easy to define and train for different fairness definitions, analogous to those that are popular for binary classification problems. 

One key distinction is between unsupervised and supervised fairness metrics:
for example, consider the task of ranking restaurants for college students who prefer cheaper restaurants, and suppose we wish to be fair to French vs Mexican restaurants. Our proposed unsupervised \emph{statistical parity} constraint would require that the model be equally likely to (i) rank a French restaurant above a Mexican restaurant, and (ii) rank a Mexican restaurant above a French restaurant. 
%\emph{exposure}-based fairness criteria would require that French and Mexican restaurants be equally visible near the top of the ranking (a.k.a.\ disparate treatment \cite{SinghJ18}). %, or that the ratio of French and Mexican restaurants appearing at the top be the same as the relative proportion of French and  Mexican restaurants in the locality (a.k.a.\ equal attention \cite{Biega+18}). 
In contrast, our proposed supervised \emph{equal opportunity} constraint would require that the model be equally likely to (i) rank a cheap French restaurant above an expensive Mexican restaurant, and (ii) rank a cheap Mexican restaurant above an expensive French restaurant. %We will give similar pairwise-based definitions for regression models, and will also show how our proposed framework extends to continuous sensitive attributes like age.

% treated the same as cheap Mexican restaurants, i.e.\ the probability that a cheap French restaurant is ranked above an expensive Mexican restaurant, is the same as the probability that a cheap Mexican restaurant is ranked above an expensive French restaurant. That is, we show how to achieve rankings that are fair in the sense that \emph{high-quality} members of each group are likely to be ordered correctly \textit{across} and \textit{within} each group. 

Like some recent work on fair ranking \citep{SIRpairwise:2019, KallusZ19}, we draw inspiration from the standard learning-to-rank strategy~\citep{learningToRankBook}: we reduce the ranking problem to that of learning a binary classifier to predict the relative ordering of \emph{pairs} of examples.
%
%Our work extends beyond the prior fair ranking work in that (i) we show how to formulate a broader set of statistical fairness metrics in terms of such pairwise comparisons, (ii) we provide analogous fairness definitions for regression, (iii) we handle continuous protected attributes not just discrete, (iv) we show how to effectively train ranking and regression models to satisfy the proposed fairness metrics by applying state-of-the-art constrained optimization algorithms. 
%
This reduction of ranking to binary classification enables us to formulate a broad set of statistical fairness metrics, inspired by analogues in the binary classification setting, in terms of pairwise comparisons. The same general idea can actually be applied more broadly: in addition to group-based fairness in the ranking setting, we show that the same overall approach can also be applied to (i) the regression setting, or (ii) the use of continuous protected attributes instead of discrete groups. In all three of these cases, we show how to effectively train ranking models or regression models to satisfy the proposed fairness metrics, by applying state-of-the-art constrained optimization algorithms.

\section{Ranking Pairwise Fairness Metrics}\label{sec:ranking}

We begin by considering a standard ranking set-up~\citep{learningToRankBook}: we're given a sample $S$ of queries drawn \iid from an underlying distribution $\mathcal{D}$, where each query is a set of candidates to be ranked, and each candidate is represented by an associated feature vector $x \in \mathcal{X}$ and label $y \in \mathcal{Y}$. The label space can be, for example, $\mathcal{Y} = \{0,1\}$ (\eg for click data: $y=1$ if a result was clicked by a user, $y=0$ otherwise), $\mathcal{Y} = \R$ (each result has an associated quality rating), or $\mathcal{Y} = \N$ (the labels are a ground truth ranking). We adopt the convention that higher labels should be ranked closer to the top. Any of these choices of label space $\mathcal{Y}$ induce a partial ordering on examples, for which all candidates belonging to the \emph{same} query are totally ordered, and any two candidates $(x,y)$ and $(x',y')$ belonging to \emph{different} queries are incomparable.
%
%, where $y > y'$ is as in the ``Pairwise Fairness'' section. % \secref{pf}. 

%\section{Pairwise Fairness}\label{sec:pf}
%\input{figures/tab-matrix}

Suppose that we have a set of $K$ protected groups $G_1,\dots,G_K$ partitioning the space of examples $\mathcal{X} \times \mathcal{Y}$ such that every example belongs to exactly one group. We define the \emph{group-dependent pairwise accuracy} $\pacc{G_i}{G_j}$ as the accuracy of a ranking function $f: \cX \rightarrow \R$ on those pairs for which the labeled ``better'' example belongs to group $G_i$, and the labeled ``worse'' example belongs to group $G_j$. That is:
\begin{eqnarray}
  \lefteqn{
  \pacc{G_i}{G_j} \defeq}
  \label{eq:pairAccuracy}  
  \\
  &  \probability ( f(x) > f(x') \mid
  y > y', (x,y) \in G_i, (x',y') \in G_j ),
\nonumber
\end{eqnarray}
where $(x,y)$ and $(x',y')$ are drawn \iid from the distribution of examples, restricted to the appropriate protected groups. Notice that this definition implicitly forces us to construct pairs only from examples belonging to the same query, since $y$ and $y'$ are not comparable if they belong to different queries---however, the probability is taken over all such pairs, across \emph{all} queries. Given $K$ groups, one can compute the $K \times K$ matrix of all possible $K^2$ group-dependent pairwise accuracies. %(see Table \ref{tab:matrix}).  
One can also measure how each group performs \emph{on average}:
\begin{align}
  \pacc{G_i}{:} \defeq& \probability ( f(x) > f(x') \mid y > y', (x,y) \in G_i ) \label{eq:pairAccuracyRowMarginal} \\
  \pacc{:}{G_i} \defeq& \probability ( f(x) > f(x') \mid y > y', (x',y') \in G_i ). \label{eq:pairAccuracyColMarginal}
\end{align}
The accuracy in \eqref{pairAccuracyRowMarginal} is averaged over all pairs for which the $G_i$ example was labeled as ``better,'' and the ``worse" example is from any group, including  $G_i$. Similarly, \eqref{pairAccuracyColMarginal} is the  accuracy averaged over all pairs where the $G_i$ example should not have been preferred. Lastly, the overall pairwise accuracy  $\probability (f(x) > f(x') \mid y > y')$ is simply the standard AUC. 
Next, we use the pairwise accuracies to define \emph{supervised} pairwise fairness goals and \emph{unsupervised} fairness notions.

% Maya says: cut for space, I don't it's adding enough value for how much space it takes up. 
%\input{figures/tab-matrix}

%These group-dependent pairwise metrics will enable us to construct ranking and regression analogues of the equalized odds fairness metric~\citep{Hardt:2016} and the equal accuracy metric~\citep{Cotter:2019b}. The correspondence between pairwise comparisons and pairwise binary classification enables us to leverage machinery developed for training fairer binary classifiers to train ranking and regression models with constraints on the proposed pairwise fairness metrics. 

%and have a deep connection with the recently-proposed \emph{pinned AUC metric} of \citet{Dixon:2018}. 

\subsection{Pairwise Equal Opportunity} We construct a \emph{pairwise equal opportunity}  analogue of the \emph{equal opportunity} metric~\citep{Hardt:2016}: 
\begin{equation}
  \pacc{G_i}{G_j} = \kappa, \mbox{ for some } \kappa \in [0,1], \mbox{ for all } i, j \label{eq:pairEqualOpp}
\end{equation}
\emph{Equal opportunity} for binary classifiers~\citep{Hardt:2016} requires positively-labeled examples to be equally likely to be predicted positively regardless of protected group membership. Similarly, this \emph{pairwise equal opportunity} for ranking problems requires \emph{pairs} to be equally-likely to be ranked correctly regardless of the protected group membership of both members of the pair. By symmetry, we could equally well consider $\pacc{G_i}{G_j}$ to be a true positive rate or a true negative rate, so there is no distinction between ``equal opportunity'' and ``equal odds'' in the ranking setting, when all of the pairwise accuracies are constrained equivalently.

\emph{Pairwise equal opportunity} can be relaxed either by requiring all pairwise accuracies 
%(i) to be above some desired lower bound (\eg $\pacc{G_i}{G_j} \ge 0.7$ for all $i \ne j$), 
(i) to only be within some quantity of each other (\eg $\max_{i \ne j} \pacc{G_i}{G_j} - \min_{i \ne j} \pacc{G_i}{G_j} \le 0.1$), or (ii) only requiring the minimum pairwise accuracy $\pacc{G_i}{G_j}$ to be as big as possible (\ie $\maximize \min_{i \ne j} \pacc{G_i}{G_j}$), in the style of robust optimization~\egcite{Chen:2017}. We will later show %in the ``Training for Pairwise Fairness'' section %\secref{algorithm} 
how models can be efficiently trained subject to both these types of pairwise fairness constraints using existing algorithms.

%\subsection{In-Group vs. Cross-Group Comparison (Abstract Version)}
 
%In the previous section, we proposed a pairwise equal opportunity metric that treated all $\pacc{G_i}{G_j}$s equivalently. We have observed, however, that in many cases diagonal elements (with $i=j$) behave very differently from the off-diagonals ($i \ne j$). There can be a variety of reasons for this: (i) it may be easier to compare an element with another element of the same group, than to an element of a different group (perhaps because certain features are more or less important depending on the group); (ii) conversely, groups could be sufficiently distinguished that across-group comparisons would be easier than within-group comparisons; or (iii) label noise could be group-dependent, resulting in the differences in noise levels being magnified for within-group comparisons.
%
%To address these (and related) issues, we suggest a refinement of \eqref{pairEqualOpp} that requires the diagonal elements to be equal, and likewise for the off-diagonals, but does not require the two sets of elements to be equal to each other:
%
%\begin{align}
  %
%  \pacc{G_i}{G_j} =& \kappa, \mbox{ for some } \kappa \in [0,1], \mbox{ for all } i \ne j \label{eq:splitPairEqualOpp} \\
  %
%  \pacc{G_i}{G_i} =& \kappa_{d}, \mbox{ for some } \kappa_{d} \in [0,1], \mbox{ for all } i \notag
  %
%\end{align}
%
%The second requirement---on the diagonal elements---can be interpreted a version of the ``equal accuracy'' metric of \TODO{cite}. As in \secref{ranking}, these strict equality constraints can be relaxed in various ways.

\subsection{Within-Group vs. Cross-Group Comparison} We have observed that labels for within-group comparisons ($i = j$) are sometimes more accurate and consistent across raters than labels for cross-group comparisons ($i \ne j$) can be noisier and less consistent. This especially arises when the labels are coming from experts that are more comfortable with rating candidates from certain groups.  For example, consider a video ranking system where group $i$ is sports videos and group $j$ is cooking shows. If our experts can choose which videos they rate (as in most consumer recommendation systems with feedback), sports experts are likely to rate sports videos and do so accurately, cooking experts are likely to rate cooking shows and do so accurately, but on average we may not get as accurate ratings on pairs with a sports and cooking video.  

Thus one may wish to \emph{separately} constrain \emph{cross-group pairwise equal opportunity}:
\begin{align}
  \pacc{G_i}{G_j} =& \kappa, \mbox{ for some } \kappa \in [0,1] \mbox{ for all } i \ne j. \label{eq:crossGroupPairEqualOpp} 
\end{align}
and \emph{within-group pairwise equal accuracy}:~\\[-15pt]
\begin{align}
  \pacc{G_i}{G_i} =& \kappa', \mbox{ for some } \kappa' \in [0,1], \mbox{ for all } i. \label{eq:pairEqualAccuracy}
\end{align}
In certain applications, particularly those in which cross-group comparisons are rare or do not occur, we might want to constrain \emph{only} pairwise equal accuracy \eqref{pairEqualAccuracy}. For example, we might want a music ranking system to be equally accurate at ranking jazz as it is at ranking country music, but avoid trying to constrain cross-group ranking accuracy because we may not have confidence in cross-group ratings.

\subsection{Marginal Equal Opportunity} The previous pairwise equal opportunity proposals are defined in terms of the $K^2$ group-dependent pairwise accuracies. This may be too fine-grained, either for statistical significance reasons, or because the fine-grained constraints might be infeasible. To address this, we propose a looser \emph{marginal pairwise equal opportunity} criterion that asks for parity for each group averaged over the other groups:
\begin{equation}
  \pacc{G_i}{:} = \kappa \mbox{ for some } \kappa \in [0,1], \mbox{ for } i = 1, \ldots, K.  \label{eq:marginalPairEqualOpp}
\end{equation}

\subsection{Statistical Parity} 
Our pairwise setup can also be used to define unsupervised fairness metrics.  For any $i \ne j$, we define \emph{pairwise statistical parity}  as:
\begin{align}
\probability \left( f(x) > f(x') \mid (x,y) \in G_i, (x',y') \in G_j \right) = \kappa.   \label{eq:pairParity}
\end{align}
A pairwise statistical parity constraint requires that if two candidates are compared from different groups, then on average each group has an equal chance of being top-ranked. This constraint completely ignores the training labels, but that may be useful when groups are so different that any comparison is too \emph{apples-to-oranges} to be legitimate, or if raters are not expert enough to make useful cross-group comparisons. %For example, if the labels come from raters or users who are experts at rating examples from one group, but not the other. For example, if one group is jazz music and another group is celtic music, and the raters are either jazz or celtic music fans, we do not expect meaningful cross-group ratings.

\section{Regression Pairwise Fairness Metrics}\label{sec:regression}

Consider the standard regression setting in which $f: \cX \rightarrow \mathcal{Y}$ attempts to predict a regression label for each example. For most of the following proposed regression fairness metrics, we treat higher scores as more (or less) desirable, and we seek to control how often each group gets higher scores. This asymmetric perspective is applicable if the scores confer a benefit, such as regression models that estimate credit scores or  admission to college, or if the model scores dictate a penalty to be avoided, such as getting stopped by police. This asymmetry assumption that getting higher scores is either preferred (or not-preferred) is analogous to the binary classification case where a positive label is assumed to confer some benefit.

We again propose defining metrics on pairs of examples.  This is not a ranking problem, so there are no queries---instead, given a training set of $N$ examples, we compute pairwise metrics over all $N^2$ pairs. One can sample a \textit{random subset} of pairs if $N^2$ is too large. 

\subsection{Regression Equal Opportunity} One can compute and constrain the pairwise equal opportunity metrics as in \eqrefsss{pairEqualOpp}{crossGroupPairEqualOpp}{pairEqualAccuracy}{marginalPairEqualOpp} for regression models. For example, restricting \eqref{crossGroupPairEqualOpp} constrains the model to be equally likely for all groups $G_i$ and $G_j$ to assign a higher score to group $i$ examples over group $j$ examples, if the group $i$ example's label is higher. 

\subsection{Regression Equal Accuracy} Promoting \emph{pairwise equal accuracy} as in \eqref{pairEqualAccuracy} for regression requires that, for every group, the model should be equally faithful to the pairwise ranking of any two within-group examples. This is especially useful if the regression labels of different groups originate from different communities, and have different labeling distributions. For example, suppose that all jazz music examples are rated by jazz lovers who only give 4-5 star ratings, but all classical music examples are rated by critics who give a range of 1-5 star ratings, with 5 being rare. Simply minimizing MSE alone might cause the model training to over-focus on the classical music score examples, since the classical errors are likely to be larger and hence affect the MSE more. 

\subsection{Regression Statistical Parity}
For regression, the pairwise statistical parity condition described in \eqref{pairParity} requires, ``Given two randomly drawn examples from two different groups, they are equally likely to have the higher score.'' One sufficient condition to guarantee pairwise statistical parity is to require the distribution of outputs $f(X)$ for a random input $X$ to be the same for each of the protected groups. 
%Note that this is weaker than the statistical parity definition of \citep{Agarwal:2019} which requires matching the output distributions for different protected groups in a point-wise manner.
This condition can be enforced approximately by histogram matching the output distributions for different protected groups \egcite{Agarwal:2019}. 

\subsection{Regression Symmetric Equal Accuracy} For regression problems where each group's goal is to be accurate (rather than to score high or low), one can define symmetric pairwise fairness metrics as well, for example, the symmetric pair accuracy for group as $G_i$ is $\pacc{G_i}{:} + \pacc{:}{G_i}$, and one might constrain these accuracies to be the same across groups. %This strategy is different than trying to equate the squared loss across groups, because this metric is purely ordinal and more similar to the Kendall-Tau metric~\citep{learningToRankBook}.  

\section{Continuous Protected Features}\label{sec:continuous}
Suppose we have a continuous or ordered protected feature $Z$; \eg we may wish to constrain for fairness with respect to age, income, seniority, etc. The proposed pairwise fairness notions extend nicely to this setting by constructing the pairs based on the ordering of the protected feature, rather on protected group membership. Specifically, we change \eqref{pairAccuracy} to the following \emph{continuous attribute pairwise accuracies}:  
\begin{align}
    \cacc{>} \defeq \probability ( f(x) > f(x') \mid y > y', z > z' ),\label{eq:pairContinuous} \\
    \cacc{<} \defeq \probability ( f(x) > f(x') \mid y > y', z < z' ),
\end{align}
where $z$ is the protected feature value for $(x,y)$ and $z'$ is the protected feature value for $(x', y')$. For example, if the protected feature $Z$ measures \emph{height}, then $\cacc{>}$ measures the accuracy of the model when comparing pairs where the candidate who is taller should receive a higher score.

The previously proposed pairwise fairness constraints for discrete groups have analogous definitions in this setting by replacing \eqref{pairAccuracy} with
\eqref{pairContinuous}. Pairwise equal opportunity becomes \begin{equation}
\cacc{>}  = \cacc{<}.\label{eq:pairContinuousEqOpp}
\end{equation}
This requires, for example, that the model be equally accurate when the taller or shorter candidate should be higher ranked.\footnote{Similar to the pairwise ranking metrics, % \secref{ranking}, 
$\cacc{<}$ is the true negative rate for pairs $(x,y), (x',y')$ where $z > z'$, and by symmetry, $\cacc{<}$ is also equal to the true positive rate for pairs where $z < z'$:
\begin{align*}
    \cacc{<} &= \probability ( f(x) < f(x') \mid y < y', z > z' ) \\
    &= \probability ( f(x) > f(x') \mid y > y', z < z' )
    ~~=~ TPR_{z < z'}. 
\end{align*}
Therefore, \eqref{pairContinuousEqOpp} equates both the TPR and the TNR for both sets of pairs, and specifies both equalized odds and equal opportunity.}
 %As detailed in \appref{TPR}, $\cacc{<}$ is a true negative rate for pairs $(x,y), (x',y')$ where $z > z'$, and thus \eqref{pairContinuousEqOpp} equates both the true positive rates and the true negative rates for both sets of pairs, and specifies both equalized odds and equal opportunity.

% MAYA and SERENA Commented this out because the value wasn't clear and it seemed to make things more complex.  
%The analogous pairwise equal accuracy condition becomes
%\begin{equation*}
%\probability \left\{ f(x) > f(x') \mid y > y', z = z' \right\} = \kappa,
%\end{equation*}
%that is, that the model be equally accurate for all hunger levels at ranking pairs of equally-hungry candidates.

%If higher $Z$ corresponds to a more misfortunate candidate (\eg a hungrier child), than Rawls Principle of Indifference can be implemented by an inequality constraint...

% optimization problems and notes on algorithms used here

\section{Training for Pairwise Fairness}
\label{sec:algorithm}
We show how one can use the pairwise fairness definitions to specify a training objective, and how to optimize these objectives. We formulate the training problem for ranking and cross-group equal opportunity, but the formulation and algorithms can be applied to any of the pairwise metrics.

\subsubsection{Proposed Formulations}
\label{sec:formulations}
Let $\pacc{G_i}{G_j}(f)$ be defined by \eqref{pairAccuracy} for a ranking model $f: \mathcal{X} \rightarrow \R$. Let $AUC(f)$ be the overall pairwise accuracy. % (equivalently the area under the ROC curve). 
Let $\cF$ be the class of models we are interested in. We formulate training with fairness goals as a \textit{constrained optimization} with an allowed slack $\epsilon$:
\begin{align}
%\textbf{Constrained opt.: }\\
 \max_{f \in \cF}  AUC(f)  \hspace{2.5cm}
  \nonumber\\
  \suchthat \;\; \pacc{G_i}{G_j}(f) - \pacc{G_k}{G_l}(f) \leq \epsilon \;\;
\forall i \ne j, k \ne l.
\label{eq:constrained}
\end{align}
Or one can pose the \textit{robust optimization} problem:
\begin{align}
% \textbf{Robust opt.: } \\
  \max_{f \in \cF, \,\xi}   \xi   \hspace{2.5cm}
  \nonumber\\
   \suchthat \;\; \xi \leq AUC(f), \xi \leq \pacc{G_i}{G_j}(f) \;\; \forall i \ne j. \label{eq:robust}
\end{align}

%Let $\cG$ denote the set of pairs $(i, j)$ that we wish to compare. For example, for cross-group equal opportunity, $\cG$ would be set of off-diagonal entries of $A$. 

%Since it could be difficult to satisfy constraints forcing all cross-group pairs to have the same pairwise accuracy, we propose two alternate ways to relax the fairness requirement. These relaxations will then allow us to formulate the learning problem in terms of binary rate constraints and re-use existing constrained learning frameworks for binary classification to solve the resulting optimization problem.\\

%\begin{equation}
%\label{eq:constrained}
%\textbf{Constrained opt.: } \max_{f \in \cF}  AUC(f) \;\; \suchthat \;\; \pacc{G_i}{G_j}(f) - \pacc{G_k}{G_l}(f) \leq \epsilon \;\;
%\forall i \ne j, k \ne l.
%\end{equation}

%\begin{equation} \label{eq:robust}
% \textbf{Robust opt.: } \max_{f \in \cF, \xi \geq 0}   \xi \;\; %\suchthat \;\; \xi \leq AUC(f), \xi \leq \pacc{G_i}{G_j}(f) \;\; %\forall i \ne j. %(i, j) \in \cG
%\end{equation}

%These two problem formulations can be varied to handle the other pairwise fairness metrics we have proposed. 
For regression problems, we replace AUC with MSE. 

%use the constrained optimization formulation with the maximum AUC objective replaced with minimum mean squared error (MSE) as the objective. 

%We do not use the robust optimization approach for regression tasks as the squared error is not necessarily comparable with the regression pairwise metrics.

%To apply the formulations to the marginal equal opportunity criterion, we would define similar constraints on row-based averages $\pacc{G_i}{:}(f)$ for all $i$ (see \eqref{pairAccuracyRowMarginal}). To apply the formulations to continuous protected features, we would define constraints on  the protected pairwise accuracies $\cacc{>}(f)$ and $\cacc{<}(f)$ (see \eqref{pairContinuous}). 

\subsubsection{Optimization Algorithms}
Both the constrained and robust optimization formulations can be written in terms of \emph{rate constraints}~\citep{Goh:2016} on score differences. For example, we can re-write each pairwise accuracy term as a positive prediction rate on a subset of pairs:
\begin{equation*}
\pacc{G_i}{G_j}(f) \,=\, \expectation\left[\indicator_{f(x) - f(x') > 0} ~\big|~ ((x, y), (x', y')) \in \cS_{ij} \right],
\end{equation*}
where $\indicator$ is the usual indicator function and $\cS_{ij} \,=\, \{((x, y), (x', y')) %\in (\cX \times \cY)^2 
\,|\, y > y', (x,y) \in G_i, (x',y') \in G_j\}$. This enables us to adopt algorithms for binary fairness constraints to solve the optimization problems in \eqrefs{constrained}{robust}.

In fact, \emph{all} of the objective and constraint functions that we have considered can be handled out-of-the-box by the proxy-Lagrangian framework of \citet{Cotter:2019,Cotter:2019b}. Like other constrained optimization approaches~\citep{Agarwal:2018, Kearns:2018}, this framework learns a \textit{stochastic model} that is supported on a finite set of functions in $\cF$.  
The high-level idea is to set up a min-max game, where one player minimizes over the model parameters, and the other player maximizes over a weighting $\lambda$ on the constraint functions. 
% For example, for formulation \eqref{constrained} with two protected groups, denoting $\Delta_{ij}(f) = \pacc{G_{i}}{G_{j}}(f) - \pacc{G_{j}}{G_{i}}(f)$, the min-max game is given by:
% \[
% \min_{\lambda_+, \lambda_- \geq 0}\max_{f \in \cF}\,  AUC(f)
% \,+\,
% \lambda_+ (\Delta_{12}(f) - \epsilon)
% \,+\,
% \lambda_- (\Delta_{21}(f) - \epsilon).
%  \]
\citet{Cotter:2019} use a no-regret optimization strategy for minimization over the model parameters, and a swap-regret optimization strategy for maximization over $\lambda$, with the indicators $\indicator$  replaced with hinge-based surrogates for the first player \emph{only}. They prove that, under certain assumptions, their optimizers converge to a stochastic model that satisfies the specified constraints in expectation. In the Appendix, we present more details about the optimization approach and re-state their theoretical result for our setting.%\footnote{The appendix can be found in the full version of the paper: \url{https://arxiv.org/pdf/1906.05330.pdf}}

\section{Related Work} \label{sec:relatedWork}
We review related work that we build upon in fair classification, and then related work on the problems addressed here: fair ranking, fair regression, and handling continuous protected attributes. 

%We also point out similarities/differences with recently proposed AUC-based metrics for fairness.
%\subsection{Related Work: Fair Ranking} \label{sec:relatedOther}

\subsubsection{Fair Classification}
Many statistical fairness metrics for binary classification can be written in terms of \emph{rate constraints}, that is, constraints on the classifier's positive (or negative) prediction rate for different groups~\citep{Goh:2016,Narasimhan:2018,Cotter:2019,Cotter:2019b}. For example, the goal of \emph{demographic parity}~\citep{dwork:2012} is to ensure that the classifier's positive prediction rate is the same across all protected groups. Similarly, the \emph{equal opportunity} metric~\citep{Hardt:2016} requires that true positive rates
% which can be expressed in terms of the classifier's positive prediction rates on positively-labeled examples,
should be equal across all protected groups. Many other statistical fairness metrics can be expressed in terms of rates, \eg \emph{equal accuracy}, \emph{no worse off} and \emph{no lost benefits}~\citep{Cotter:2019b}. Constraints on these fairness metrics can be added to the training objective for a binary classifier, then solved using constrained optimization algorithms or relaxations thereof~\citep{Goh:2016,Zafar:2017,Donini:2018,Agarwal:2018,Cotter:2019,Cotter:2019b}. Here, we extend this work to train ranking models and regression models with pairwise fairness constraints.

\subsubsection{Fair Ranking} % Most prior fair ranking work addresses \emph{unsupervised} exposure metrics ~\egcite{Zehlike:2017,Zehlike:2018,Celis:2018,Biega+18}, rather than \emph{supervised} fairness no
A majority of the previous work on fair ranking has focused on list-wise definitions for fairness  that depend on the entire list of results for a given query \egcite{Zehlike:2017,Celis:2018,Biega+18,SinghJ18,Zehlike:2018,SinghJ19}. These include both \textit{unsupervised} criteria that require  the average exposure near the top of the ranked list to be equal for different groups
\egcite{SinghJ18, Celis:2018, Zehlike:2018},
and \textit{supervised} criteria that require the average exposure for a group to be proportional to the average relevance of that group's results to the query \citep{Biega+18, SinghJ18, SinghJ19}. Of these, some provide post-processing algorithms for re-ranking a given ranking \citep{Biega+18, Celis:2018, SinghJ18, SinghJ19}, while others, like us, learn a ranking model from scratch \citep{Zehlike:2018, SinghJ19}.

% relevance/quality of the ranked examples, but for the special case of re-ranking a given ranking. In a recent work, they extend this to a more general listwise ranking setting \cite{SinghJ19}. \citet{Biega+18} also addresses the re-ranking setting, given \emph{fairness constraints} on the  frequency of exposure of different groups of items between any two ranking positions. 
%Pinned AUC  and weighted pinned AUC~\citep{Borkan:2019} are fairness metrics that can be written as a linear combination of within-group and cross-group pairwise accuracies $\pacc{G_i}{G_j}$~\citep{Borkan:2019}, and a de-biasing re-sampling strategy was proposed to improve pinned AUC~\citep{Dixon:2018}. %; see Section \ref{sec:} for a more detailed discussion. 

\subsubsection{Pairwise Fairness} %There have been two very recent works on pairwise fairness for ranking \citep{SIRpairwise:2019, KallusZ19}.
\citet{SIRpairwise:2019} propose ranking pairwise fairness definitions equivalent to those we give in \eqrefss{pairAccuracy}{pairAccuracyRowMarginal}{pairAccuracyColMarginal}. Their work focuses on ranking and on categorical protected groups, whereas we generalize these ideas to capture a wider variety of different statistical fairness notions, and generalize to regression and continuous protected features. 

The training methodology is also very different.
\citet{SIRpairwise:2019} propose adding a fixed regularization term to the training objective that measures the \emph{correlation} between the residual between a clicked and unclicked item and the group memberships of the items. In contrast, we enable explicitly specifying any desired pairwise fairness constraints, and then directly enforce the desired pairwise fairness criterion using constrained optimization. Their approach is parameter-free, but only because it does not give the user any way to control the trade-off between fairness vs. accuracy. 

%Maya 5pm day of submission:This is redundant and we are short on space
%Our approach \emph{(i)} enables the user to choose a point on the Pareto front by specifying their choice of fairness constraint, and \emph{(ii)} enables us to leverage existing work on constrained optimization for machine learning, including the state-of-the-art algorithms such as \citet{Cotter:2019}, and the provably improved-generalization strategy of \citet{Cotter:2019c}.
 
Second, \citeauthor{SIRpairwise:2019} consider only two protected groups, whereas we enable the user to constrain any number of groups, with the constrained optimization algorithm automatically determining how much each group must be penalized in order to satisfy the fairness constraints. A straightforward extension of the fixed regularization approach of \citeauthor{SIRpairwise:2019} to multiple groups would have no hyperparameters to specify how much to weight each group. One could introduce separate weighting hyperparameters to weight each group's penalty, but then they would need to be tuned manually. The approach we propose does this tuning \emph{automatically} to achieve the desired fairness constraints. 
 
% \citet{Cotter:2019b} have noted that such constraint formulations are easier for practitioners to specify and test. % Moreover, previous work comparing relaxing to correlation vs directly solving the constrained optimization problem has shown substantial wins for the more direct approach we take here~\citep{Goh:2016}. 
Finally, there are major experimental differences to \citeauthor{SIRpairwise:2019}: they provide an in-depth case study of one real-world recommendation problem, whereas we provide a broad set of experiments on public and real-world data illustrating the effectiveness on both ranking and regression problems, for categorical or continuous protected attributes.

%Another recent work by \citet{KallusZ19} also provide pairwise fairness metrics based on AUC for bipartite ranking problems. However, they only consider categorical groups, whereas we also handle regression problems and continuous protected attributes. Further, their methodology is a post-processing approach that fits a monotonic transformation to an existing ranking model  to optimize the specified metrics. In contrast, we provide a more flexible approach that enables optimizing the entire model by including the desired metrics as constrains during training.

In other recent works, \citet{KallusZ19} provide pairwise fairness metrics based on AUC for bipartite ranking problems, and \cite{Kuhlman+19} provide metrics analogous to our pairwise equal opportunity and statistical parity criteria for ranking. Both these works only consider categorical groups, whereas we also handle regression problems and continuous protected attributes. \citet{KallusZ19} additionally propose a post-processing optimization approach  to optimize the specified metrics by fitting a monotone transformation to an existing ranking model. In contrast, we provide a more flexible approach that enables optimizing the entire model by including explicit constraints during training.

\subsubsection{Pinned AUC}
Pinned AUC is a fairness metric introduced by \citet{Dixon:2018}. % and the weighted pinned AUC by the authors represents another. 
With two protected groups, pinned AUC works by resampling the data such that each of the two groups make up 50\% of the data, and then calculating the ROC AUC on the resampled dataset. Based on the well-known equivalence between ROC AUC and average pairwise accuracy, \citet{Borkan:2019} demonstrate that pinned AUC, as well as their proposed weighted pinned AUC metric, can be decomposed as a linear combination of within-group and cross-group pairwise accuracies. In other words, both pinned AUC and weighted pinned AUC can be written as linear combinations of different pairwise accuracies $\pacc{G_i}{G_j}$ in \eqref{pairAccuracy}.
%
%Thus pinned AUC is one possible weighting of the pairwise accuracy matrix.  We propose a broader set of fairness criteria for ranking and regression that are expressed as constraints on entries of the pairwise accuracy matrix. Our provided optimization algorithms automatically find the right weighting of the matrix entries to satisfy the specified fairness criteria (see the ``Training for Pairwise Fairness'' section  %\secref{weighted-pairs}  for details). 
%
In our experiments, we compare against (a version of) the sampling-based approach of \citet{Dixon:2018}. %This serves as a representative baseline that optimizes a fixed weighting of the pairwise accuracy matrix.

%\subsection{Related Work: Binary Classifiers} %\label{sec:relatedBinary}

% Andy says: Discuss multi-head models~\egcite{Beutel:2017}?
% Maya says: what aspect is relevant here? So far we're restricting
% our discussion of related work on binary classifier fairness to stuff we use - is the multi-head models relevant to what we do here? Or you want to contrast with it?

%\subsection{Related Work: Fair Regression} %\label{sec:relatedRegression}

\subsubsection{Fair Regression} Defining fairness metrics in a regression setting is a challenging task, and has been studied for many years in the context of standardized testing~\egcite{HunterSchmidt:1976}. \citet{Komiyama:2018} consider the unfairness of a regressor in terms of the correlation between the output and a protected attribute. \citet{perez2017fair}  regularize to minimize the Hilbert-Schmidt independence between the protected features and model output. These definitions have the ``flavor'' of statistical parity, in that they attempt to remove information about the protected feature from the model's predictions. Here, we focus more on \emph{supervised} fairness notions. 

\citet{Berk:2017} propose regularizing linear regression models for the notion of fairness corresponding to the principle that \emph{similar individuals receive similar outcomes}~\citep{dwork:2012}. Their definitions focus on enforcing similar squared error, which fundamentally differs from our definitions in that we assume each group would prefer higher scores, not necessarily more accurate scores. 

\citet{Agarwal:2019} propose a \emph{bounded group loss}  definition which requires that the regression error be within an allowable limit for each  group. In contrast, our pairwise equal opportunity definitions for regression do not rely on a specific regression loss, but instead are based on the ordering induced by the regression model within and across groups. %\citeauthor{Agarwal:2019} 
% 
%They also propose a statistical parity definition based on matching the output distributions for different groups.
%\subsection{Related Work: Continuous Protected Features}\label{sec:relatedContinuous}

\subsubsection{Continuous Protected Features}  Most prior work in machine learning fairness has assumed categorical protected groups, in some cases extending those tools to continuous features by bucketing~\citep{Kearns:2018}. Fine-grained buckets raise statistical significance challenges, and coarse-grained buckets may raise unfairness issues due to how the lines between bins are drawn, and the lack of distinctions made between element within each bin. \citet{Raff:2018} considered continuous protected features in their tree-growing criterion that addresses fairness. \citet{Kearns:2018} focused on statistical parity-type constraints for continuous protected features for classification. \citet{Komiyama:2018} controlled the correlation of the model output with protected variables (which may be continuous). \citet{Mary:2019} propose a fairness criterion for continuous attributes based on the R{\'e}nyi maximum correlation coefficient. \emph{Counterfactual fairness}~\citep{Kusner:2017, Pearl:2016} requires that changing a protected attribute, while holding causally unrelated attributes constant, should not change the model output distribution, but this does not directly address issues with ranking fairness. 

%, and certain protected attributes that are often treated as categorical may be better treated as continuous. For example, in the Communities and Crime dataset~\citep{crime:2002}, each community is described by what percentage of its population is from each race, but this feature is typically bucketized into categorical racial groups. We show in \secref{continuous} that a proposed pairwise fairness notion can be applied to real-valued protected attributes without bucketization.

%\input{figures/tab-simulated}
%%%%%%%%%%%%%%%%%%%%%
\begin{figure*}
\begin{subfigure}[t]{0.25\textwidth}
\centering
\begin{tabular}{cc}
  & \hspace{1.2cm}\textbf{Negative} \\
  \rotatebox[origin=c]{90}{\textbf{Positive}} \hspace{-10pt} &
  \begin{tabular}{c|cc}
    & \textbf{$G_0$} & \textbf{$G_1$} \\
    \hline
    \textbf{$G_0$} &    0.941 &    0.980 \\
    \textbf{$G_1$} &    0.705 &    0.894 \\
    \hline
  \end{tabular}
  \\
\end{tabular}
\caption{Sim./Unconstrained}
\end{subfigure}
\begin{subfigure}[t]{0.25\textwidth}
\centering
\begin{tabular}{cc}
  & \hspace{1.2cm}\textbf{Negative} \\
  \rotatebox[origin=c]{90}{\textbf{Positive}} \hspace{-10pt} &
  \begin{tabular}{c|ccc|c}
    & \textbf{$G_0$} & \textbf{$G_1$} \\
    \hline
    \textbf{$G_0$} &    0.854 &   { 0.900} \\
    \textbf{$G_1$} &    {0.907} &    0.927 \\
    \hline
  \end{tabular}
  \\
\end{tabular}
\caption{Sim./Constrained}
\end{subfigure}
\begin{subfigure}[t]{0.24\textwidth}
\centering
\begin{tabular}{ccc}
    \textbf{$G_0$} & \textbf{$G_1$} & \textbf{$G_2$} \\
    \hline
    %
    %  0.779 &  0.759 &  0.779\\
    % 0.710 &  0.700  &  0.728\\
    % 0.714 & 0.699 & 0.729\\
    0.713 & 0.703   & 0.729\\
    \hline
\end{tabular}
\caption{Sim./Constrained}
\end{subfigure}
\begin{subfigure}[t]{0.24\textwidth}
\centering
\begin{tabular}{ccc}
     \textbf{$G_0$} & \textbf{$G_1$} & \textbf{$G_2$} \\
    \hline
    %
    %   0.930 &  0.879 & 0.885\\
    % 0.881 & 0.877 & 0.949\\
    0.882 & 0.876 & 0.949\\
    \hline
\end{tabular}
\caption{Sim./Robust}
\end{subfigure}
% \begin{subfigure}[t]{0.18\textwidth}
% \centering
% \begin{tabular}{cc}
%     %
%     \textbf{C} & \textbf{NC} \\
%     %
%     \hline
%     %
%      0.767 &    0.706\\
%     \hline
%     %
%     %
% \end{tabular}
% \caption{Business/Uncons.}
% \end{subfigure}
% %
% \begin{subfigure}[t]{0.17\textwidth}
% \centering
% \begin{tabular}{cc}
%     %
%      \textbf{C} & \textbf{NC} \\
%     %
%     \hline
%     %
%       {0.715} &    {0.712}\\
%     \hline
%     %
%     %
% \end{tabular}
% \caption{Business/Constr.}
% \end{subfigure}
% \begin{subfigure}[t]{0.18\textwidth}
% \centering
% \begin{tabular}{cc}
%     %
%      \textbf{C} & \textbf{NC} \\
%     %
%     \hline
%     %
%       {0.804} &    0.733\\
%     \hline
%     %
%     %
% \end{tabular}
% \caption{Business/Robust}
% \end{subfigure}
%
\\[3pt]
\begin{subfigure}[t]{0.32\textwidth}
\centering
\begin{tabular}{cc}
  &\hspace{0.9cm} \textbf{Non-toxic} \\
  \rotatebox[origin=c]{90}{\textbf{Toxic}}\hspace{-10pt} &
  \begin{tabular}{c|ccc|c}
    & \textbf{Other} & \textbf{Gay} \\
    \hline
    \textbf{Other} &  0.973 &  0.882 \\
    \textbf{Gay}   &  0.986 &  0.937 \\
    \hline
  \end{tabular}
  \\
\end{tabular}
\caption{Wiki/Unconstrained}
\end{subfigure}
\begin{subfigure}[t]{0.32\textwidth}
\centering
\begin{tabular}{cc}
  &\hspace{0.9cm} \textbf{Non-toxic} \\
  \rotatebox[origin=c]{90}{\textbf{Toxic}}\hspace{-10pt} &
  \begin{tabular}{c|ccc|c}
    & \textbf{Other} & \textbf{Gay} \\
    \hline
    \textbf{Other} &  0.971 &  {0.953} \\
    \textbf{Gay}   &  {0.967} &  0.945 \\
    \hline
  \end{tabular}
  \\
\end{tabular}
\caption{Wiki/Debiased}
\end{subfigure}
\begin{subfigure}[t]{0.32\textwidth}
\centering
\begin{tabular}{cc}
  &\hspace{0.9cm} \textbf{Non-toxic} \\
  \rotatebox[origin=c]{90}{\textbf{Toxic}}\hspace{-10pt} &
  \begin{tabular}{c|ccc|c}
    & \textbf{Other} & \textbf{Gay} \\
    \hline
    \textbf{Other} &  0.962 &  {0.943} \\
    \textbf{Gay}   &  {0.953} &  0.922 \\
    \hline
  \end{tabular}
  \\
\end{tabular}
\caption{Wiki/Constrained}
\end{subfigure}
\vspace{-4pt}
%\caption{Test pairwise accuracy matrices for  Simulated ranking (a)--(b),  row-based matrix averages $\pacc{chain}{:}$  and $\pacc{non-chain}{:}$ for Business matching (c)--(e), and pairwise accuracy matrices for Wiki Talk Pages ranking (f)--(h).}
\caption{Test pairwise accuracy matrices for  Simulated ranking with 2 groups (a)--(b),  row-based matrix averages $\pacc{0}{:}$, $\pacc{1}{:}$ and $\pacc{2}{:}$ for Simulated ranking with 3 groups (c)--(d), and pairwise accuracy matrices for Wiki Talk Pages ranking (e)--(g).}
\label{fig:matrix-sim-law}
\vspace{-1pt}
\end{figure*}
%%%%%%%%%%%%%%%%%%%%%

\begin{table*}[t]
\caption{Test AUC (higher is better) with test pairwise fairness violations (in parentheses). For fairness violations, we report $|\pacc{G_0}{G_1} - \pacc{G_1}{G_0}|$ when imposing cross-group constraints, %$\max\{|\pacc{G_0}{G_1} - \pacc{G_1}{G_0}|,~ |\pacc{G_0}{G_0} - \pacc{G_1}{G_1}|\}$ when imposing both cross-group and in-group constraints, 
$|\pacc{G_0}{:} - \pacc{G_1}{:}|$ for marginal constraints,  and $|\cacc{>} - \cacc{<}|$ for %the constraint on 
continuous protected attributes.
\hlgt{Blue} indicates strictly best between \cite{SIRpairwise:2019} and constrained optimization.
}
\vspace{-4pt}
    \centering
    % \begin{tabular}{c|c|c|c|c|c}
    %\begin{tabular}{l|c|c|c|c}
    \begin{tabular}{lcccccccc}
        \hline
        \textbf{Data} & 
        \textbf{Groups} &
        \textbf{Uncons.} &
        \textbf{Debiased} &
        \textbf{S \& J} &
        \textbf{K \& Z} &
        \textbf{B et al.} &
        \textbf{Constr.} &  
        \textbf{Robust}
        \\         
        \hline
Sim. & 0/1 &
\textbf{0.92} ~(0.28) & 
\textbf{0.92} ~(0.28) &
0.88 ~(0.14) &
0.91 ~(0.12) &
0.84 ~(0.04) &
\hlgt{0.86 ~(\textbf{0.01})} & 
0.86 ~(0.02) 
\\%\hline
% Sim. CG \& IG & 0/1 & \textbf{0.92} ~(0.28) &\textbf{0.92} ~(0.28) & 0.75 ~({0.05}) &{0.89} ~(\textbf{0.05})  \\%\hline
Busns. & C/NC &
\textbf{0.70} ~(0.06) & 
\textbf{0.70} ~(0.06) &
--&
0.66 ~(\textbf{0.00}) &
0.69 ~(0.05) &
0.68 ~(\textbf{0.00}) &
0.68 ~(0.07) \\%\hline
Wiki & Term `Gay' & 
\textbf{0.97} ~(0.10) &
\textbf{0.97} ~(\textbf{0.01}) &
-- &
\textbf{0.97} ~(0.04) &
0.95 (\textbf{0.01}) &
\hlgt{0.96 ~(\textbf{0.01})} &
0.94 ~(0.02)\\%\hline
W3C & Gender &
{0.53} ~(0.96) &
{0.54} ~(0.90) &
0.37 ~(0.85) &
0.45 ~(0.65) &
\hlgt{\textbf{0.55} ~(\textbf{0.09})}  &
{0.54} ~(0.10) & 
{0.54} ~(0.14) \\ %\hline
Crime & Race \%& 
\textbf{0.93} ~(0.18) &
-- &
-- &
-- &
0.91 (0.10) &
0.81 ~(\textbf{0.04}) &
{0.86} ~(\textbf{0.04}) \\
\hline
\end{tabular}
\label{tab:results-summary}
\vspace{-10pt}
\end{table*}
%\input{figures/tab-summary-regression}
%%%%%%%%%%%%%%%%%%%%%%%%%%
%%%
\begin{table*}
\begin{minipage}[t]{0.65\textwidth}
    \centering
    % \begin{tabular}{c|c|c|c|c|c|c|c}
    \begin{tabular}{lcccc}
        \hline
        \textbf{Dataset} & 
        \textbf{Prot. Group} &
        % \textbf{Problem} &
        %\textbf{Constraint Type} &  
        \textbf{Unconstrained} &
        \textbf{Beutel et al.} &
        \textbf{Constrained}  \\
        \hline
        Law &
        \textit{Gender} &
        %Cross-group &
        \textbf{0.142} ~(0.30) &
        0.167 ~(0.06) &
        \hlgt{0.143 ~(\textbf{0.02})}
        \\
        %
        % \multirow{2}{*}{Crime} &
        Crime &
        % \multirow{2}{*}{\textit{Race} (binary)} &
        % \textit{Race} &
        % \textbf{0.021}~(0.518) &
        % -- &
        % 0.098~(\textbf{0.019}) &
        % -- 
        % %
        % %
        % &
        % &
        % &
        % &
        % Cross-group  \& In-group &
        % \textbf{0.021} ~(0.483) &
        % 0.047 ~(\textbf{0.061}) 
        % \\
        \textit{Race \%} &
       % Cont.\ Attr.\ &
        \textbf{0.021} ~(0.33) &
        0.033 ~(\textbf{0.02}) &
        0.028 ~({0.03}) 
        \\
        \hline
    \end{tabular}
    %\label{tab:results-summary-regression}
    % \vspace{-14pt}
\end{minipage}
\begin{minipage}[t]{0.34\textwidth}
\centering
\begin{tabular}{cc}
  & \hspace{1cm} \textbf{Low} \\
  \rotatebox[origin=c]{90}{\textbf{High}} \hspace{-10pt} &
  \begin{tabular}{c|ccc|c}
    & \textbf{Male} & \textbf{Female} \\
    \hline
    \textbf{Male}   &  0.652 &   {0.647} \\
    \textbf{Female} &  {0.666} &   0.655 \\
    \hline
  \end{tabular}
  \\
\end{tabular}
\end{minipage}
\caption{Left: Regression test MSE (lower is better) and pairwise fairness violation (in parenthesis), with \hlgt{blue} indicating strictly best between last two columns. Right: Test pairwise accuracy matrix for \textit{constrained} optimization on Law School dataset. }
\label{tab:results-summary-regression}
\vspace{-10pt}
\end{table*}
%%%
%%%%%%%%%%%%%%%%%%%%%%%%%%

\section{Experiments}\label{sec:experiments}
We illustrate our proposals %for improving pairwise fairness metrics on 
on five ranking problems and two regression problems. 
%We implemented the proposed constrained and robust optimization methods described in \secref{algorithm} in TensorFlow using the  \emph{proxy Lagrangian} optimization in the open-source TensorFlow Constrained Optimization toolbox~\citep{Cotter:2019, Cotter:2019b}.  Our code will be made publicly available. 
%\subsection{Experimental Details}
We implement the constrained and robust optimization methods % described % in \secref{algorithm}
using the open-source Tensorflow constrained optimization toolbox of \cite{Cotter:2019, Cotter:2019b}.  
The datasets used are split randomly into training, validation and test sets in the ratio 1/2:1/4:1/4, with the validation set used to tune the relevant hyperparameters.  
%In each case, the dataset is split into training, validation and test sets in the ratio 1/2:1/4:1/4. 
% \if 0
% \footnote{We use Adam for gradient updates, and use hinge loss based proxy constraints. 
% For the Wiki Talk Pages dataset and the Law School dataset, we use minibatches of 100 stochastic gradients to better handle the large number of pairs to be enumerated. For all other datasets, we compute full gradients. In each case, the dataset is split into training, validation and test sets in the ratio 1/2:1/4:1/4, with the validation set used to tune the learning rate,  the number of training epochs for the unconstrained optimization methods, and for the post-processing shrinking step in the proxy-Lagrangian solver. Our code will be made publicly available.} 
% \fi 
For datasets with queries, we evaluate all  metrics for individual queries and report the average across queries. %The constrained and robust optimization methods learn a stochastic model, and the  metrics reported for these methods are
For stochastic models, we report expectations over random draws of the scoring function $f$ from the stochastic model.\footnote{Code available at: \url{https://github.com/google-research/google-research/tree/master/pairwise_fairness}}

%%%
\subsection{Pairwise Fairness for Ranking}
%%%
We detail the comparisons and ranking problems. 

\subsubsection{Comparisons} We compare against: (1) an adaptation of the \textit{debiasing} scheme of \citet{Dixon:2018} that optimizes a weighted pairwise accuracy, with the weights chosen to balance the relative  label  proportions  within  each group;  (2) the recent non-pairwise ranking fairness approach by \citet{SinghJ18} that re-ranks the scores of an unconstrained ranking  model to satisfy a disparate impact constraint; (3) the post-processing pairwise fairness method of \citet{KallusZ19} that fits a monotone transform to an unconstrained model; and (4) the fixed regularization pairwise approach of \citet{SIRpairwise:2019} that like us incorporates the fairness goal into the model training. See Appendix for more details.

\subsubsection{Simulated Ranking Data}\label{app:simulation}
For this toy ranking task with two features, there are 5,000 queries, and each query has 11 candidates. For each query, we uniformly randomly pick one of the 11 candidates to have a positive label $y=+1$ and the other 10 candidates receive a negative label $y=-1$, and we randomly assign each candidate's protected attribute $z$ \iid from a $Bernoulli(0.1)$ distribution. Then we generate two features simulated to score how well the candidate matches the query, from a Gaussian distribution 
$\mathcal{N}(\mu_{y, z},\, \Sigma_{y, z})$, where
$
\mu_{-1,0} = [-1,1],~
\mu_{-1,1} = [-2,-1],~
\mu_{+1,0} = [1, 0],~
\mu_{+1,1} = [-1.5, 0.75],
$
$
\Sigma_{-1,0} \,=\, 
\Sigma_{-1,1} \,=\, 
\Sigma_{+1,0} \,=\, \mathbf{I}_2$
and
$\Sigma_{+1,1} \,=\, 0.5\,\mathbf{I}_2
$.
% \fi

% \begin{figure*}
%     \centering
%     \includegraphics[scale=0.37]{figures/synthetic_data_decision_boundaries.png}
%     \vspace{-10pt}
%     \caption{Plot of learned hyperplanes on simulated ranking data. For constrained and robust optimization, we plot the hyperplane that is assigned the highest probability in the support of the learned stochastic model.  
%     %in the model's support.
%     }
%     \label{fig:decision-boundaries-simulated}
%     %\vspace{-5pt}
% \end{figure*}

We train linear ranking functions $f:\mathbb{R}^2 \rightarrow \mathbb{R}$
and impose a \textit{cross-group equal opportunity} with constrained optimization by constraining  $|\pacc{0}{1} - \pacc{1}{0}| \le 0.01$. For the robust optimization, we implement this goal by maximizing $\min\{\pacc{0}{1},\, \pacc{1}{0},\, AUC\}$. We also train an unconstrained model that optimizes AUC.
% For the 2nd experiment we seek to enforce \textit{cross-group equal opportunity and in-group equal accuracy} by constraining both $|\pacc{0}{1} - \pacc{1}{0}| \le 0.01$ and $|\pacc{0}{0} - \pacc{1}{1}| \le 0.01$. For the robust optimization, we implement this goal by  maximizing   $\min\{\pacc{0}{1},\, \pacc{1}{0}\}$ + $\min\{\pacc{0}{0},\, \pacc{1}{1}\}$.
%
\tabref{results-summary} gives the test ranking accuracy, and the test pairwise fairness  violations, measured as $|\pacc{0}{1} - \pacc{1}{0}|$. 
%for the first fairness criterion and as $\max\{|\pacc{0}{1} - \pacc{1}{0}|,\, |\pacc{0}{0} - \pacc{1}{1}\}$ for the second criterion. 
 %Debiasing is not helpful in this case. 
 Only constrained optimization achieves the  fairness goal, with robust optimization coming a close second. 
%For the 2nd experiment, the robust optimization does better with the second fairness goal.
\figref{matrix-sim-law}(a)--(b) shows the $2 \times 2$ pairwise accuracy matrices. Constrained optimization satisfies the fairness constraint by lowering $\pacc{0}{1}$ and improving $\pacc{1}{0}$.  
%In the Appendix, 

We also generate a second dataset with 3 groups, where the first two groups follow the same distribution as groups 0 and 1 above, and the third group examples are drawn from a Gaussian distribution $\mathcal{N}(\mu_{y, 2},\, \Sigma_{y, 2})$ where $\mu_{-1,2} = [-1,1], \mu_{+1,2}=[1.5, 0.5]$, and $\Sigma_{-1,2} =
\Sigma_{+1,2} = \mathbf{I}_2$. We use the same number of queries and candidates as above, and assign the protected attribute $z$ to $0, 1$, or $2$ with probabilities $0.45, 0.1$, and $0.45$ respectively. We
impose the \textit{marginal equal opportunity} fairness goal on this dataset in two different ways: \textit{(i)} constraining  $\max_{i \ne j} |\pacc{i}{:} - \pacc{j}{:}| \le 0.01$ with constrained optimization, and \textit{(ii)} optimizing $\min\{AUC, \pacc{0}{:}, \pacc{1}{:}, \pacc{2}{:}\}$ with robust optimization. 
We show each group's row-marginal test accuracies in \figref{matrix-sim-law}(c)--(d). While robust optimization maximizes the minimum of the three marginals, constrained optimization yields a lower difference  between the marginals (and does so at the cost of lower accuracies for the three groups). This is consistent with the two optimization problem set-ups: you get what you ask for.

We provide further results and an additional experiment with an \textit{in-group} equal opportunity criterion in the appendix.

\subsubsection{Business Matching} This is a proprietary  dataset from a large internet services company of ranked pairs of relevant and irrelevant businesses for different queries, for a total of 17,069 pairs. How well a query matches a candidate is represented by 41 features. We consider two protected groups, \emph{chain} (C) businesses and \emph{not chain} (NC) businesses. We define a candidate as a member of the \emph{chain} group %$G_{chain}$ 
if its query is seeking a chain business and the candidate is a chain business. We define a candidate as a member of the \emph{non-chain} group %$G_{non-chain}$ 
if its query is not seeking a chain business and the candidate is a non-chain business.  A candidate does not belong to either group if it is chain and the query is non-chain-seeking, or vice-versa.  

%For this experiment,
We experiment with imposing a marginal equal opportunity constraint:
$
|\pacc{chain}{:} \,-\, \pacc{non-chain}{:}| \,\leq\, 0.01.
$
This requires the model to be roughly as accurate at correctly matching chains as it at matching non-chains. With robust optimization, we maximize $\min\{\pacc{chain}{:},\, \pacc{non-chain}{:},\, AUC\}$.  All methods trained a two-layer neural network model with 10 hidden nodes.  
As seen in \tabref{results-summary}, compared to the unconstrained approach, constrained optimization yields very low fairness violation, while only being marginally worse on the test AUC. The post-processing approach of \citep{KallusZ19} also achieves a similar fairness metric, but with a lower  AUC. \cite{SinghJ18} failed to produce feasible
solutions for this dataset, we believe because there were very few pairs per query.

\subsubsection{Wiki Talk Page Comments} This public dataset contains 127,820 comments from Wikipedia Talk Pages labeled with whether or not they are toxic (\ie contain ``rude, disrespectful or unreasonable'' content~\citep{Dixon:2018}). This is a dataset where debiased weighting has been effective in learning fair, unbiased classification models~\citep{Dixon:2018}. We consider the task of learning a ranking function that ranks  comments that are labeled toxic higher than the comments that are labeled non-toxic, in order to help the model's users identify toxic comments. We consider the protected attribute defined by whether the term `gay' appears in the comment. This is one of the many identity terms that \citet{Dixon:2018} consider in their work.  Among comments that have the term `gay', 55\% are labeled toxic, whereas among comments that do not have the term `gay', only 9\% are labeled toxic. We learn a convolutional neural network model with the same architecture used in \citet{Dixon:2018}. 

We consider a cross-group equal opportunity criterion. We impose
$|\pacc{Other}{Gay} - \pacc{Gay}{Other}| \le 0.01$ with constrained optimization and maximize $\min\{\pacc{Other}{Gay},\, \pacc{Gay}{Other}, AUC\}$ with robust optimization. The results are shown in \tabref{results-summary} and \figref{matrix-sim-law}(e)-(g). Among the cross-group errors, the unconstrained model is more likely to incorrectly rank a non-toxic comment with the term `gay' over a toxic comment without the term. By balancing  the label proportions, debiased weighting reduces the fairness violation considerably. Constrained optimization  yields even lower fairness violation (0.010 vs.\ 0.014), but at the cost of a slightly lower test AUC. 
\cite{SinghJ18} could not be  applied to this dataset  as it did not have the required query-candidate structure.
% \if 0
% \fi
% \input{figures/tab-wiki-toxicity}
% \input{figures/tab-experts-ranking}
% \input{figures/tab-regression}

% \input{figures/tab-wiki-toxicity}

%%%
%%%
\subsubsection{W3C Experts Search}
%%%
We also evaluate our methods on the W3C Experts dataset, previously used to study disparate exposure in ranking~\citep{Zehlike:2018}. This  is a
subset of the TREC 2005 enterprise track data, and consists of 48 topics and 200 candidates per topic, with each candidate labeled as an expert or non-expert for the topic. The task is to rank the candidates based on their expertise on a topic, using a corpus of mailing lists from the World Wide Web Consortium (W3C).  This is an application where the unconstrained algorithm does better for the minority protected group.  We use the same features as \citet{Zehlike:2018} to represent
how well each topic matches each candidate; this includes a set of five aggregate features derived from word counts and tf-idf scores, and the gender protected attribute.

For this task, we learn a linear model and impose a cross-group equal opportunity constraint: $|\pacc{Female}{Male} - \pacc{Male}{Female}| \le 0.01$. For robust optimization, we maximize $\min\{\pacc{Female}{Male}$, $\pacc{Male}{Female},\, AUC\}$.
As seen in \tabref{results-summary}, 
%and the group pairwise accuracies are shown in \figref{matrix-experts}, 
the unconstrained ranking model incurs a huge fairness violation. This is because the unconstrained model treats gender as a strong signal of expertise, and often ranks female candidates over male candidates. 
% \if 0
% As a result, $\pacc{Female}{Male}$ is close to 100\%, while $\pacc{Male}{Female}$ is close to 0. 
% \fi
Not only do the constrained and robust optimization methods achieve significantly lower fairness violations, they also happen to produce higher test metrics due to the constraints acting as regularizers and reducing overfitting. On this task, \cite{SIRpairwise:2019} achieves the lowest fairness violation and the highest AUC.

 The method of \cite{SinghJ18} performs poorly because the LP that it solves per query turns out to be infeasible for most queries in this dataset. Thus, to run this baseline, we extended their approach to have a per-query slack in their disparate impact constraints. This required a large slack for some queries, hurting the overall performance.

%\subsection{Continuous Protected Attributes}
\subsubsection{Communities and Crime (Continuous Groups)} We next handle a \textit{continuous protected attribute} in a ranking problem. We use the  \textit{Communities and Crime} dataset  from UCI~\citep{UCI} 
which contains 1,994 communities in the United States described by 140 features, and the per capita crime rate for each community. As in prior work~\citep{Cotter:2019}, we label the communities with a crime rate above the $70$th percentile as `high crime' and the others as `low crime', and consider the task of learning a ranking function that ranks high crime communities above the low crime communities.  We treat the percentage of black population in a community as a continuous protected attribute. 

We learn a linear ranking function, with the protected attribute included as a feature. We do not compare to debiasing, and the methods of \citep{SinghJ18} and \citep{KallusZ19}, as they do not apply to continuous protected attributes. 
Adopting the continuous attribute equal opportunity criterion, we impose the constraint $|\cacc{<} - \cacc{>}| \leq 0.01$. We extend \citep{SIRpairwise:2019} to optimize this pairwise metric.  \tabref{results-summary} shows the constrained and robust optimization methods reduce the fairness violation by more than half, at the cost of a lower test AUC.

\subsection{Pairwise Fairness for Regression}
We next present experiments on two regression problems.

We extend the set-up of \citet{SIRpairwise:2019} to also handle our proposed regression pairwise metrics, and compare to that.
We do not use robust optimization as the squared error is not necessarily comparable with the regression pairwise metrics. The results are shown in \tabref{results-summary-regression}. 

\textit{Law School:} This dataset~\citep{Wightman:1998} contains details of 27,234 law school students, and we predict the undergraduate GPA for a student from the student's LSAT score, family income, full-time status, race, gender and the law school cluster the student belongs to, with gender as the protected attribute. We impose a cross-group equal opportunity constraint: $|\pacc{Female}{Male} - \pacc{Male}{Female}| \leq 0.01$. %\tabref{results-summary-regression} shows
The constrained optimization approach successfully massively reduces the fairness violation compared to the unconstrained MSE-optimizing model, at only a small increase in MSE. It also performs strictly better than \citet{SIRpairwise:2019}.
% \if 0
% The group pairwise accuracies are the Law School regression problem are shown in \figref{law}.
% \fi

\textit{Communities and Crime:} This dataset has continuous labels for the per capita crime rate for a community. Once again, we treat the percentage of black population in a community as a \textit{continuous protected attribute} and impose a continuous attribute equal opportunity constraint: $|\cacc{>} - \cacc{<}| \leq 0.01$. 
%The results are shown in \tabref{results-summary-regression}.  
% and \figref{crime-regression}. 
The constrained approach yields a huge reduction in fairness violation, though at the cost of an increase in MSE.

\section{Conclusions}
We showed that pairwise fairness metrics can be intuitively defined to handle supervised and unsupervised notions of fairness, for ranking and regression, and for discrete and continuous protected attributes. We also showed how pairwise fairness metrics can be incorporated into training using state-of-the-art constrained optimization solvers.  %One could  additionally bring in side information and weight the pairs differently, e.g.\ based on the order in which the candidates were presented to the  user when labeled.  

Experimentally, the different methods compared often produced different trade-offs between AUC (or MSE) and fairness, making it hard to judge one as strictly better than others. However, we showed that the proposed constrained optimization approach is the most flexible and direct of the strategies considered, and 
is very effective for achieving low pairwise fairness violations. The closest competitor to our proposals is the approach of \citet{SIRpairwise:2019}, but out of the 4 cases in which one of these two methods strictly performed better (indicated in blue), ours was the best in 3 of the 4. Lastly, \citet{KallusZ19}, being a post-processing method, is more restricted, and did not perform as well as the proposed approach that directly trains a model from scratch.

The key way one specifies pairwise fairness metrics is by the selection of which pairs to consider. Here, we focused on within-group and cross-group candidates. 
%One could  additionally bring in side information and weight the pairs differently, e.g.\ based on the order in which the candidates were presented to the  user when labeled.  
One could also bring in side information or condition on other features. For example, in the ranking setting, we might have side information about the presentation order that candidates for each query were shown to users when labeled, and this position information could be used to either select or weight candidate pairs. In the regression setting, we could assign weights to example pairs based on their label differences.

%However, constraints must be carefully specified to reflect the actual goals, otherwise equality across groups can produce worse overall performance.
%The de-biasing heuristic of \citep{Dixon:2018} performs well  on the Wiki Talk Pages dataset, but is not very useful for the other problems. \citet{SinghJ18} seeks to satisfy a non-pairwise fairness criterion and performs poorly  on the proposed pairwise metrics. 

 %In the regression setting, we could select or weight example pairs based on the differences in labels within pairs.

%We compared to the de-biasing heuristic of \citep{Dixon:2018}, and were able to obtain good results for the Wiki Talk Pages dataset studied by them, but did not find the heuristic to perform well for other problems.  

\bibliographystyle{plainnat}
\bibliography{main}

%\input{appendix}
%%%%%%%%%%%%%%%
% \if 0
\newpage
\onecolumn
\appendix
\begin{center}
\LARGE	
\textbf{Appendix}
\end{center}

\begin{table}[H] 

\centering

\caption{Example group-dependent pairwise accuracy matrix of $K=3$ groups, and the frequency-weighted row mean and column mean for the restaurant ranking example in the introduction.}

\label{tab:matrix}
\small
\begin{tabular}{cc}
  & \textbf{Expensive} \\[5pt]
  \hspace{-15pt}\rotatebox[origin=c]{90}{\textbf{Cheap}}
 \hspace{-15pt}	  
   &
  \begin{tabular}{c|ccc|c}
    & \textbf{French} & \textbf{Mexican} & \textbf{Chinese} & \textbf{Row Mean} \\
    \hline
    \textbf{French} & \textit{0.76} & 0.68 & 0.56 & 0.70 \\
    \textbf{Mexican} & 0.61 & \textit{0.62} & 0.55 & 0.60 \\
    \textbf{Chinese} & 0.62 & 0.61 & \textit{0.56} & 0.60 \\
    \hline
    \textbf{Col.\ Mean} & 0.69 & 0.65 & 0.56 & \textit{0.65}
  \end{tabular}
  \\
\end{tabular}

\end{table}

\section{Proxy-Lagrangian Optimization}
For completeness, we provide details of the proxy-Lagrangian optimization approach for solving the constrained optimization in \eqref{constrained}. For ease of exposition, we consider a setting with two protected groups, i.e. $K=2$. % and will enforce a cross-group equal opportunity constraints. 
Denoting 
$\Delta_{ij}(f) = \pacc{G_{i}}{G_{j}}(f) - \pacc{G_{j}}{G_{i}}(f)$, we can restate \eqref{constrained} for two protected groups as:
\begin{align}
 \max_{\theta \in \Theta}  AUC(f_\theta)  ~~~~
  \suchthat \;\;
  \Delta_{01}(f_\theta) \leq \epsilon, \;\;
  \Delta_{10}(f_\theta) \leq \epsilon,
\label{eq:constrained-2}
\end{align}
where the ranking model $f_\theta$ is parameterized by $\theta$ and $\Theta$ is a set of model parameters we wish to optimize over.

Since the pairwise metrics are non-continuous in the model parameters (due to the presence of indicator functions $\indicator$), the approach of \cite{Cotter:2019} requires us to choose a surrogate for the AUC objective that is concave in $\theta$ and satisfies:
\[
\widetilde{AUC}(f_\theta)  ~\leq~ AUC(f_\theta),
\]
and surrogates for the constraint differences $\Delta_{ij}$ that are convex in $\theta$ and satisfy:
\[
\tilde{\Delta}_{ij}(f_\theta) ~\geq~ {\Delta}_{ij}(f_\theta).
\]
In our experiments, we construct these surrogates by replacing the indicator functions $\indicator$ in the pairwise metrics with hinge-based upper/lower bounds. \cite{Cotter:2019} then introduce Lagrange multipliers or weighting parameters for the objective and constraints $\lambda$ and
defines two proxy-Lagrangian functions:
\[
\cL_\theta(f_\theta, \lambda) ~\defeq~
\lambda_1 \widetilde{AUC}(f_\theta) \,+\,
\lambda_2 \left(\tilde{\Delta}_{01}(f_\theta) - \epsilon\right) \,+\,
\lambda_3 \left(\tilde{\Delta}_{10}(f_\theta) - \epsilon\right);
\]
\[
\cL_\lambda(f_\theta, \lambda) ~\defeq~
\lambda_2 \left({\Delta}_{01}(f_\theta) - \epsilon\right) \,+\,
\lambda_3 \left({\Delta}_{10}(f_\theta) - \epsilon\right),
\]
where $\lambda \in  \Delta^3$  is chosen from the 3-dimensional simplex. The goal is to maximize $\cL$ over the model parameters, and minimize $\cL_\lambda$ over $\lambda$, resulting in a two-player game between a $\theta$-player and a $\lambda$-player. 

Note that the $\theta$-player's objective $\cL_\theta$ is defined using the surrogates $\widetilde{AUC}$ and  $\tilde{\Delta}_{ij}$s (this is needed for optimization over model parameters), while the second player's objective $\cL_\lambda$ is defined using the  original pairwise metrics (as the optimization over $\lambda$ does not need the indicators to be relaxed). 

The proxy-Lagrangian approach uses a no-regret optimization strategy with step-size $\eta_\theta$ to optimize $\cL_\theta$ over model parameters $\theta$, and a swap-regret optimization strategy with step-size $\eta_\lambda$ to optimize $\cL_\lambda$ over $\lambda$. 
See Algorithm 2 of \citet{Cotter:2019} for specific details of this iterative procedure. The result of $T$ steps of this optimization is a stochastic model $\bar{\mu}$ supported on $T$ models, which under certain assumptions, is guaranteed to satisfy the specified cross-group constraints. We adapt and re-state the convergence guarantee of \citet{Cotter:2019}  for our setting:

\begin{theorem}[\citet{Cotter:2019}]
Let $\Theta$ be a compact convex set. For a given choice of step-sizes $\eta_\theta$ and $\eta_\lambda$, let $\theta_1, \ldots, \theta_T$ and $\lambda_1, \ldots, \lambda_T$ be the iterates returned by the proxy-Lagrangian optimization algorithm (Algorithm 2 of \cite{Cotter:2019}) after $T$ iterations.
Let $\bar{\mu}$ be a stochastic model with  equal probability on each of $f_{\theta_1}, \ldots, f_{\theta_T}$ and let $\bar{\lambda} = \frac{1}{T}\sum_{t=1}^T \lambda_t$. Then there exists choices of step sizes $\eta_\theta$ and $\eta_\lambda$ for which $\bar{\mu}$ thus obtained satisfies with probability $\geq 1 - \delta$ (over draws of stochastic gradients by the algorithm):
\[
\expectation_{f \sim \bar{\mu}}\left[\widetilde{AUC}(f)\right] ~\geq~ \max_{f\,:\,\tilde{\Delta}_{01}(f) \leq \epsilon,\, \tilde{\Delta}_{10}(f) \leq \epsilon}\widetilde{AUC}(f) ~-~ \tilde{\mathcal{O}}\left(\frac{1}{\sqrt{T}}\right)
\]
and
\[
\expectation_{f \sim \bar{\mu}}[{\Delta}_{01}(f)] ~\leq ~ \epsilon \,+\, \tilde{\mathcal{O}}\left(\frac{1}{\bar{\lambda}_1\sqrt{T}}\right);~~~~
\expectation_{f \sim {\mu}}[{\Delta}_{10}(f)] ~\leq ~ \epsilon \,+\, \tilde{\mathcal{O}}\left(\frac{1}{\bar{\lambda}_1\sqrt{T}}\right),
\]
where $\mathcal{O}$  only hides logarithmic factors.
\end{theorem}

Under some additional assumptions (that there exists model parameters that strictly satisfy the specified constraints with a margin), \citet{Cotter:2019} show that for a large $T$, the term  $\bar{\lambda}_1$ in the above guarantee is essentially lower-bounded by a constant, and the returned stochastic model is guaranteed to closely satisfy the desired pairwise fairness constraints in expectation.

\begin{figure*}[t]
    \centering
    \includegraphics[scale=0.37]{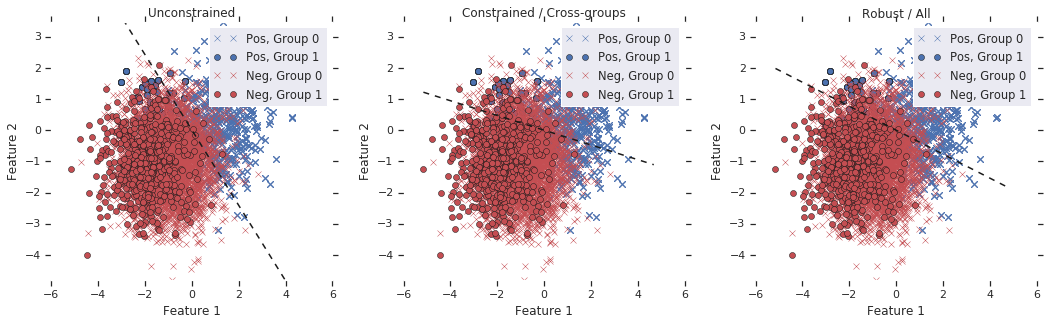}
    \vspace{-10pt}
    \caption{Plot of learned hyperplanes on simulated ranking data with 2 groups. For constrained and robust optimization, we plot the hyperplane that is assigned the highest probability in the support of the learned stochastic model.  
    %in the model's support.
    }
    \label{fig:decision-boundaries-simulated}
    %\vspace{-5pt}
\end{figure*}

\begin{table*}[t]
\caption{Test AUC (higher is better) with test pairwise fairness violations (in parentheses) for a simulated ranking task with two groups and two sets of fairness goals: a \textit{cross-group} equal opportunity criterion and an \textit{in-group} equal opportunity criterion. For fairness violations, we report $\max\{|\pacc{G_0}{G_1} - \pacc{G_1}{G_0}|,~ |\pacc{G_0}{G_0} - \pacc{G_1}{G_1}|\}$.
}
    \centering
    % \begin{tabular}{c|c|c|c|c|c}
    %\begin{tabular}{l|c|c|c|c}
    \begin{tabular}{lcccccccc}
        \hline
        \textbf{Dataset} & 
        \textbf{Prot.\ Group} &
        \textbf{Unconstrained} &
        \textbf{Debiased} &
        \textbf{Constrained} &  
        \textbf{Robust}
        \\         
                \hline
Sim. CG \& IG & 0/1 & \textbf{0.92} ~(0.28) &\textbf{0.92} ~(0.28) & 0.75 ~({0.05}) &{0.89} ~(\textbf{0.05})  \\\hline
\end{tabular}
\label{tab:results-summary-ig}
\end{table*}
\begin{figure*}[h!]
\footnotesize
\begin{subfigure}[t]{0.32\textwidth}
\centering
\begin{tabular}{cc}
  & \hspace{1.2cm}\textbf{Negative} \\
  \rotatebox[origin=c]{90}{\textbf{Positive}} \hspace{-10pt} &
  \begin{tabular}{c|cc}
    & \textbf{$G_0$} & \textbf{$G_1$} \\
    \hline
    \textbf{$G_0$} &    0.941 &    0.980 \\
    \textbf{$G_1$} &    0.705 &    0.894 \\
    \hline
  \end{tabular}
  \\
\end{tabular}
\caption{Unconstrained}
\end{subfigure}
\begin{subfigure}[t]{0.32\textwidth}
\centering
\begin{tabular}{cc}
  & \hspace{1.2cm}\textbf{Negative} \\
  \rotatebox[origin=c]{90}{\textbf{Positive}} \hspace{-10pt}
  &
  \begin{tabular}{c|ccc|c}
    & \textbf{$G_0$} & \textbf{$G_1$} \\
    \hline
    \textbf{$G_0$} &    {0.743} &    {0.769} \\
    \textbf{$G_1$} &    {0.779} &    {0.797} \\
    \hline
  \end{tabular}
  \\
\end{tabular}
\caption{Constrained}
\end{subfigure}
\begin{subfigure}[t]{0.32\textwidth}
\centering
\begin{tabular}{cc}
  & \hspace{1.2cm}\textbf{Negative} \\
  \rotatebox[origin=c]{90}{\textbf{Positive}}
  \hspace{-10pt} &
  \begin{tabular}{c|ccc|c}
    & \textbf{$G_0$} & \textbf{$G_1$} \\
    \hline
    \textbf{$G_0$ } &    {{0.881}} &    0.930 \\
    \textbf{$G_1$} &    0.897 &    0.935 \\
    \hline
  \end{tabular}
\end{tabular}
\caption{Robust}
\end{subfigure}
\caption{Test pairwise accuracy matrix for simulated (ranking) data two groups, with both cross-group and in-group criteria.}
\label{fig:matrix-simulated}
\vspace{-5pt}
\end{figure*}

\begin{table*}[t]
\caption{Test AUC (higher is better) with test pairwise fairness violations (in parentheses) for a simulated ranking task with 3 groups and with the \textit{marginal} equal opportunity fairness goal. For fairness violations, we report $\max_{i \ne j}\,|\pacc{G_i}{:} - \pacc{G_j}{:}|$.
}
    \centering
    % \begin{tabular}{c|c|c|c|c|c}
    %\begin{tabular}{l|c|c|c|c}
    \begin{tabular}{lcccccccc}
        \hline
        \textbf{Dataset} & 
        \textbf{Prot.\ Group} &
        \textbf{Unconstrained} &
        \textbf{Constrained}  &
        \textbf{Robust}
        \\         
                \hline
Sim. Marginal & $\{0, 1, 2\}$ 
& {0.95} ~(0.28) & 
{0.72 ~({0.03})} & 
0.91 (0.07)
\\
\hline
\end{tabular}
\vspace{5pt}
\label{tab:results-marginal}
\end{table*}

\subsection{Connection to Weighted Pairwise Accuracy}
\label{sec:weighted-pairs}
As discussed in the related works, pinned AUC \citep{Dixon:2018} is a previous fairness metrics that is a weighted sum of the entries of the pairwise accuracy matrix: $\sum_{i,j} \beta_{i,j} \pacc{G_i}{G_j}(f)$. The constrained and robust optimization formulations that we propose are equivalent to maximizing the weighted pairwise accuracy for a specific, optimal set of weights $\{\beta_{i,j}\}$ (see \citet{Kearns:2018,Agarwal:2018} for relationship between constrained optimization and cost-weighted learning). The constrained optimization solver essentially finds the optimal set of weights $\{ \beta_{i,j}\}$ for the two formulations. %, which may be having the each matrix entry be within $\epsilon$ of each other \eqref{constrained}, or to maximize the minimum entry \eqref{robust}. 
This is convenient because it enables us to specify fairness criteria in terms of pairwise accuracies, rather than having to manually tune the $\{\beta_{i,j} \}$ needed to produce the pairwise accuracies we want. 

\section{Additional Experimental Details/Results}
We provide further details about the experimental set-up and present additional results.
\subsubsection{Setup} We use Adam for gradient updates. 
For the Wiki Talk Pages dataset and the Law School dataset, we compute stochastic gradients using minibatches of size 100 to better handle the large number of pairs to be enumerated. For all other datasets, we enumerate all pairs and compute full gradients. The datasets are split into training, validation and test sets in the ratio 1/2:1/4:1/4, with the validation set used to tune the learning rate, the number of training epochs for the unconstrained optimization methods, and for the post-processing shrinking step in the proxy-Lagrangian solver. We have made the code available at: \url{https://github.com/google-research/google-research/tree/master/pairwise_fairness}.

\subsubsection{Comparisons}
We give more details about the debiased-weighting baseline. This is an imitation of the \textit{debiasing} scheme of \citet{Dixon:2018} by optimizing a weighted pairwise accuracy (without any explicit constraints):
\[
\frac{1}{n_+ n_-}\sum_{i,j: y_i > y_j}\, \alpha_{z_i, y_i} \alpha_{z_j, y_j} \mathbf{1}(f(x_i) > f(x_j)),
\]
where $n_+, n_-$ are the number of positively labeled and negatively labeled training examples, and $\alpha_{z, y} > 0$ is a non-negative weight on each label and protected group are set such that the relative proportions of positive and negative examples within each protected group are balanced. Specifically, we fix $\alpha_{0,-1} = \alpha_{0,+1} = \alpha_{1,+1} = 1$ and set $\alpha_{1,-1}$ so that
$
\frac{|\{x_i \,|\, z_i = 0, y_i =-1 \}|}{|\{x_i \,|\, z_i = 0, y_i =+1 \}|}
~=~
\alpha_{1,-1}\,
\frac{|\{x_i \,|\, z_i = 1, y_i =-1 \}|}{|\{x_i \,|\, z_i = 1, y_i =+1 \}|}.$ 
This mimics \citet{Dixon:2018} where they sample additional negative documents belonging group 1, so that the relative label proportions within each group are similar.

\subsubsection{Hyperparameter Choices.} 
The unconstrained approach  for optimizing AUC (MSE) is run for 2500 iterations, with the step-size chosen from the range $\{10^{-3}, 10^{-2}, \ldots, 10\}$ to maximize AUC (or minimize MSE) on the validation set.
The proxy-Lagrangian algorithm that we use for constrained and robust optimization is also run for 2500 iterations.  
We fix the step-sizes $\eta_\theta$ and $\eta_\lambda$ for this solver  to the same value, and choose this value from  the range $\{10^{-3}, 10^{-2}, \ldots, 10\}$. We use a heuristic provided in \cite{Cotter:2019b} to pick the step-size that best trades-off between the objective and constraint violations on the validation set. The final stochastic classifier for the constrained and robust optimization methods is constructed as follows: we record 100 snapshots of the iterates at regular intervals and apply the ``shrinking'' procedure of \cite{Cotter:2019b} to construct a sparse stochastic model supported on at most $J+1$ iterates, where $J$ is the number of constraints in the optimization problem.

For ranking settings, the fixed regularization approach of \cite{SIRpairwise:2019} optimizes a sum of a hinge relaxation to the AUC and their proposed correlation-based regularizer; for regression setting, it optimizes a sum of the squared error and their proposed regularizer. We run this method for 2500 iterations and choose its step-size from the range $\{10^{-3}, 10^{-2}, \ldots, 10\}$, picking the value that yields lowest regularized objective value on the validation set. For the post-processing approach of \citet{KallusZ19}, as prescribed, we fit a monotone transform  $\phi(z) \,=\, \frac{1}{1 + \exp(-(\alpha z + \beta))}$ to the unconstrained AUC-optimizing model, and tune $\alpha$  from the range $\{0, 0.05, \ldots, 5\}$ and $\beta$ from the range $\{-1, -2, -5, -10\}$, choosing values for which the transformed scores yield lowest fairness violation on the validation set. We implement the approach of \cite{SinghJ18} by solving an LP per query to post-process and re-rank the scores from the unconstrained AUC-optimizing model to satisfy their proposed disparate impact criterion. We include a per-query slack to make the LPs feasible.

\subsubsection{Experiment Constraining All Matrix Entries}
For the simulated ranking task with two groups, we present results for a second experiment, where we seek to enforce both \textit{cross-group equal opportunity and in-group equal accuracy} criteria by constraining both $|\pacc{0}{1} - \pacc{1}{0}| \le 0.01$ and $|\pacc{0}{0} - \pacc{1}{1}| \le 0.01$. For the robust optimization, we implement this goal by  maximizing   $\min\{\pacc{0}{1},\, \pacc{1}{0}\}$ + $\min\{\pacc{0}{0},\, \pacc{1}{1}\}$.
\tabref{results-summary-ig} gives the test ranking accuracy and pairwise fairness goal violations. The pairwise fairness violations are measured as  $\max\{|\pacc{0}{1} - \pacc{1}{0}|,\, |\pacc{0}{0} - \pacc{1}{1}\}$ for this experiment.
 %Debiasing is not helpful in this case. 

 As expected the unconstrained algorithm yields the highest overall ranking objective, but incurs very
high fairness violations. The debiased weighting approach also gives a similar performance (as the relative
proportions of positives and negatives are the same in expectation for the two protected groups since the
protected group was independent of the label). Both the constrained and robust optimization approaches
significantly reduce the fairness violations. In terms of the ranking objective, constrained optimization suffers
a considerable decrease in the overall objective when constraining all entries of the accuracy matrix, whereas
robust optimization incurs only a marginal decrease in objective.

\figref{matrix-simulated} shows the $2 \times 2$ pairwise accuracy matrix for each method.  From \figref{matrix-simulated}(b), one sees that  constrained optimization satisfies the fairness constraints by lowering the accuracies for all four group-pairs.  In contrast, \figref{matrix-simulated}(c), shows robust optimization  maximizes the minimum entry in the fairness matrix. These results are consistent with the two different optimization problems: you get what you ask for. 

\figref{decision-boundaries-simulated} shows the hyperplanes %intercept 
(dashed line) for the  ranking functions learned by the different methods.
% for the 2nd experiment.  
Note that the quality of the learned ranking function depends on the slope of the hyperplane and is unaffected by its intercept.
The hyperplane learned by the unconstrained approach ranks the majority examples (the + group) well, but is not accurate at ranking the minority examples (the o group). The hyperplanes learned by the constrained and robust optimization methods work more equally well for both groups. 

\subsubsection{Additional Results for Simulated Ranking Task with 3 Groups} In 
\tabref{results-marginal}, we provide the test ranking accuracy and pairwise fairness goal violations for the simulated ranking experiment with three groups, with a \textit{marginal} equal opportunity fairness criterion.

\subsubsection{Additional \cite{SinghJ18} Comparison} We also ran comparisons with the post-processing LP-based approach of \cite{SinghJ18}, with their proposed disparate treatment criterion as the constraint. The results are shown in \tabref{results-dt}.  This approach again failed to produce feasible solutions for the Business matching dataset as it had a very small number of pairs/query, and could not be applied to the Wiki Talk Pages and Communities \& Crime datasets as they did not have the  required query-candidate structure. Because this approach seeks to satisfy a non-pairwise exposure-style fairness constraint, in doing so, it fails to perform well on our  proposed pairwise metrics.

\begin{table*}[t]
\caption{Test AUC (higher is better) with test pairwise fairness violations (in parentheses) for comparisons with the approach of \cite{SinghJ18} that enforces disparate treatment. For fairness violations, we report $|\pacc{G_0}{G_1} - \pacc{G_1}{G_0}|$.
}
    \centering
    % \begin{tabular}{c|c|c|c|c|c}
    %\begin{tabular}{l|c|c|c|c}
    \begin{tabular}{lcccccccc}
        \hline
        \textbf{Dataset} & 
        \textbf{Prot.\ Group} &
        \textbf{Unconstrained} &
        \textbf{Constrained} &  
        \textbf{S \& J (DT)}
        \\         
                \hline
Sim. CG & 0/1 & {0.92} ~(0.28) & 
{0.86 ~({0.01})} & 
0.84 (0.05)\\
W3C Experts & Gender &
{0.53} ~(0.96) &
{0.54} ~(0.10) & 
0.38 (0.92)
\\
\hline
\end{tabular}
\vspace{5pt}
\label{tab:results-dt}
\end{table*}

\begin{figure*}[h]
\footnotesize
\begin{subfigure}[t]{0.45\textwidth}
\centering
\begin{tabular}{c|ccc|c}
    & \textbf{Chain} & \textbf{Non-chain} \\
    \hline
    \textbf{Row-avg} &    0.767 &    0.706\\
    \hline
\end{tabular}
\caption{Unconstrained}
\end{subfigure}
\begin{subfigure}[t]{0.45\textwidth}
\centering
\begin{tabular}{c|ccc|c}
    & \textbf{Chain} & \textbf{Non-chain} \\
    \hline
    \textbf{Row-avg} &    0.780 &    0.719\\
    \hline
\end{tabular}
\caption{Debiased}
\end{subfigure}
\\
\begin{subfigure}[t]{0.45\textwidth}
\centering
\begin{tabular}{c|ccc|c}
    & \textbf{Chain} & \textbf{Non-chain} \\
    \hline
    \textbf{Row-avg} &    {0.715} &    {0.712}\\
    \hline
\end{tabular}
\caption{Constrained}
\end{subfigure}
\begin{subfigure}[t]{0.45\textwidth}
\centering
\begin{tabular}{c|ccc|c}
    & \textbf{Chain} & \textbf{Non-chain} \\
    \hline
    \textbf{Row-avg} &    
    {0.804} &    0.733\\
    \hline
\end{tabular}
\caption{Robust}
\end{subfigure}
\caption{Test row-based matrix averages on business match (ranking) data.}
\label{fig:business}
\end{figure*}
\begin{figure*}[h!]
\begin{subfigure}[t]{0.48\textwidth}
\centering
\begin{tabular}{cc}
  & \hspace{1.1cm}\textbf{Non-expert} \\
  \rotatebox[origin=c]{90}{\textbf{Expert}} 
  \hspace{-10pt} &
  \begin{tabular}{c|ccc|c}
    & \textbf{Male} & \textbf{Female} \\
    \hline
    \textbf{Male}   &  0.534 &   0.028 \\
    \textbf{Female} &  0.991 &   0.573 \\
    \hline
  \end{tabular}
  \\
\end{tabular}
\caption{Unconstrained}
\end{subfigure}
\begin{subfigure}[t]{0.48\textwidth}
\centering
\begin{tabular}{cc}
  & \hspace{1.1cm}\textbf{Non-expert} \\
  \rotatebox[origin=c]{90}{\textbf{Expert}} \hspace{-10pt} &
  \begin{tabular}{c|ccc|c}
    & \textbf{Male} & \textbf{Female} \\
    \hline
    \textbf{Male}   &  0.541 &   0.081 \\
    \textbf{Female} &  0.977 &   0.571 \\
    \hline
  \end{tabular}
  \\
\end{tabular}
\caption{Debiased}
\end{subfigure}
\\
\begin{subfigure}[t]{0.48\textwidth}
\centering
\begin{tabular}{cc}
  & \hspace{1.1cm}\textbf{Non-expert} \\
  \rotatebox[origin=c]{90}{\textbf{Expert}} \hspace{-10pt} &
  \begin{tabular}{c|ccc|c}
    & \textbf{Male} & \textbf{Female} \\
    \hline
    \textbf{Male}   &  0.540 &   {0.501} \\
    \textbf{Female} &  {0.601} &   0.571 \\
    \hline
  \end{tabular}
  \\
\end{tabular}
\caption{Constrained/Cross-groups}
\end{subfigure}
\begin{subfigure}[t]{0.48\textwidth}
\centering
\begin{tabular}{cc}
  & \hspace{1.1cm}\textbf{Non-expert} \\
  \rotatebox[origin=c]{90}{\textbf{Expert}} \hspace{-10pt} &
  \begin{tabular}{c|ccc|c}
    & \textbf{Male} & \textbf{Female} \\
    \hline
    Male   &  0.540 &   {0.471} \\
    Female &  {0.614} &   0.553 \\
    \hline
  \end{tabular}
  \\
\end{tabular}
\caption{Robust/Cross-groups}
\end{subfigure}
\vspace{-5pt}
\caption{Test pairwise accuracy matrix for W3C experts (ranking) data.}
\label{fig:matrix-experts}
\end{figure*}
%\input{figures/tab-regression}
%\begin{figure}[t]
%%%
\begin{figure}[h!]
\footnotesize
\begin{subfigure}[t]{0.46\textwidth}
\centering
\begin{tabular}{cc}
  & \hspace{1cm} \textbf{Low} \\
  \rotatebox[origin=c]{90}{\textbf{High}}\hspace{-10pt}  &
  \begin{tabular}{c|ccc|c}
    & \textbf{Male} & \textbf{Female} \\
    \hline
    Male   &  0.653 &   0.490 \\
    Female &  0.795 &   0.654 \\
    \hline
  \end{tabular}
  \\
\end{tabular}
\caption{Unconstrained}
\end{subfigure}
\begin{subfigure}[t]{0.46\textwidth}
\centering
\begin{tabular}{cc}
  & \hspace{1cm} \textbf{Low} \\
  \rotatebox[origin=c]{90}{\textbf{High}} \hspace{-10pt} &
  \begin{tabular}{c|ccc|c}
    & \textbf{Male} & \textbf{Female} \\
    \hline
    \textbf{Male}   &  0.652 &   {0.647} \\
    \textbf{Female} &  {0.666} &   0.655 \\
    \hline
  \end{tabular}
  \\
\end{tabular}
\caption{Constrained/Cross-groups}
\end{subfigure}
\caption{Pairwise accuracy matrix for law school (regression).}
\label{fig:law}
\end{figure}
%
%
% \begin{figure}[t]
% \footnotesize
% \begin{subfigure}[t]{0.23\textwidth}
% \begin{tabular}{cc}
%   %
%   & \hspace{1cm} \textbf{Low} \\
%   %
%   \rotatebox[origin=c]{90}{\textbf{High}}\hspace{-10pt}  &
%   %
%   \begin{tabular}{c|ccc|c}
%     %
%     & \textbf{$\cacc{>}$} & \textbf{$\cacc{<}$} \\
%     %
%     \hline
%     %
%     Other &  0.742 &  0.462 \\
%     Black &  0.944 &  0.782 \\
%     \hline
%     %
%     %
%   \end{tabular}
%   %
%   \\
%   %
% \end{tabular}
% \caption{Unconstrained}
% \end{subfigure}
% %
% \begin{subfigure}[t]{0.23\textwidth}
% \begin{tabular}{cc}
%   %
%   & \hspace{1cm} \textbf{Low} \\
%   %
%   \rotatebox[origin=c]{90}{\textbf{High}}\hspace{-10pt}  &
%   %
%   \begin{tabular}{c|ccc|c}
%     %
%     & \textbf{Other} & \textbf{Black} \\
%     %
%     \hline
%     %
%     Other &  \textbf{\textit{0.678}} &  \textbf{0.733} \\
%     Black &  \textbf{0.687} &  \textbf{\textit{0.739}} \\
%     \hline
%     %
%     %
%   \end{tabular}
%   %
%   \\
%   %
% \end{tabular}
% \caption{Constrained/All}
% \end{subfigure}
% %
% \vspace{-5pt}
% \caption{Pairwise accuracy matrix for crime (regression) data.}
% \label{fig:crime-regression}
% \vspace{-10pt}
% \end{figure}
%\end{figure}
%
%
\begin{figure}[H]
\footnotesize
\begin{subfigure}[t]{0.46\textwidth}
\centering
\begin{tabular}{cc}
    $\cacc{>}$ & $\cacc{<}$ \\
    \hline
    %
    %\textbf{Row-avg} &    
    0.904 & 0.577
    \\
    \hline
\end{tabular}
\caption{Unconstrained}
\end{subfigure}
\begin{subfigure}[t]{0.46\textwidth}
\centering
\begin{tabular}{cc}
    $\cacc{>}$ & $\cacc{<}$ \\
    \hline
    %
    %\textbf{Row-avg} &    
    {0.777} & {0.745}
    \\
    \hline
\end{tabular}
\caption{Constrained/Continuous Attr.}
\end{subfigure}
\caption{Continuous attribute pairwise accuracies for crime (regression) with \%.\ of black population as protected attribute.}
\label{fig:crime-regression}
%\vspace{10cm}
\end{figure}

\subsubsection{Pairwise Accuracy Matrices}
We also present the pairwise accuracy matrices for different methods applied 
to the Business matching task (see \figref{business}), 
to the W3C experts ranking task (see \figref{matrix-experts}), to the Law School regression task (see \figref{law}), and to the Crime regression task with continuous protected attribute (see \figref{crime-regression}). In the case of the Business matching results, notice that
robust optimization maximizes the minimum of the two marginals. Despite good marginal accuracies, robust optimization's overall accuracy is not as good because of poorer behavior on the examples that were not covered by the two groups. Debiasing produces a negligible reduction in fairness violation, but yields better row marginals than the unconstrained approach.

% \fi

\end{document}